\documentclass[letterpaper, 10 pt, conference]{ieeeconf}  
\IEEEoverridecommandlockouts
\usepackage{cite}
\usepackage{amsmath,amssymb,amsfonts}
\usepackage{algorithm,algorithmic}
\usepackage{graphicx}
\usepackage{textcomp}
\usepackage{xcolor}
\graphicspath{{images/}}
\usepackage{subcaption}
\usepackage{multirow}
\usepackage{breakurl}
\usepackage{url}

\def\BibTeX{{\rm B\kern-.05em{\sc i\kern-.025em b}\kern-.08em
    T\kern-.1667em\lower.7ex\hbox{E}\kern-.125emX}}
    
\begin{document}

\title{What Are You Looking at? Detecting Human Intention in Gaze based Human-Robot Interaction\\
}

\author{\authorblockN{1\textsuperscript{st} Lei Shi}
\authorblockA{\textit{Op3Mech} \\
\textit{University of Antwerp}\\
Antwerp, Belgium \\
lei.shi@uantwerpen.be}
\and
\authorblockN{2\textsuperscript{nd} Cosmin Copot}
\authorblockA{\textit{Op3Mech} \\
\textit{University of Antwerp}\\
Antwerp, Belgium \\
cosmin.copot@uantwerpen.be}
\and
\authorblockN{3\textsuperscript{rd} Steve Vanlanduit}
\authorblockA{\textit{Op3Mech} \\
\textit{University of Antwerp}\\
Antwerp, Belgium \\
steve.vanlanduit@uantwerpen.be}
}
\maketitle

\begin{abstract}

In gaze based Human-Robot Interaction (HRI), it is important to determine the human intention for further interaction. The gaze intention is often modelled as fixation. However, when looking at an object, it is not natural and it is difficult to maintain the gaze fixating on one point for a long time. Saccades may happen while a human is still focusing on the object. The prediction of human intention will be lost during saccades. In addition, while the human intention is on object, the gazes may be located outside of the object bounding box due to different noise sources, which would cause false negative predictions. In this work, we propose a novel approach to detect whether a human is focusing on an object in HRI application. We determine the gaze intention by comparing the similarity between the hypothetic gazes on objects and the actual gazes. We use Earth Mover's Distance (EMD) to measure the similarity and 1 Nearest Neighbour to classify which object a human is looking at. Our experimental results indicate that, compare to fixation, our method can successfully determine the human intention even during saccadic eye movements and increase the classification accuracy with noisy gaze data. We also demonstrate that, in the interaction with a robot, the proposed approach can obtain a high accuracy of object selection within successful predictions.

\end{abstract}

\begin{keywords}
Human-Robot Interaction, fixation, saccade, gaze, EMD
\end{keywords}

\section{Introduction}
\label{sec:intro}

Mobile eye tracking devices i.e. eye tracking glasses usually equip eye camera(s) for detecting pupils and world camera for capturing the image of the scene. Gaze is calculated from the pupil images and it is projected to the image of the scene, which could reveal the information of human being's visual intention. Fixation and saccade are two most common types of eye movement events. Fixation can be viewed as gaze is stably kept in a small region and saccade can be viewed as rapid eye movement\cite{holmqvist2011eye}. Fixations and saccades can be computationally classified from eye tracking signals by different approach, such as dispersion (I-DT) and velocity (I-VT) based\cite{salvucci2000identifying}, Bayesian method based\cite{santini2016bayesian} and machine learning based\cite{zemblys2018using}.

\begin{figure}[t]
\centering
\includegraphics[width=0.49\textwidth]{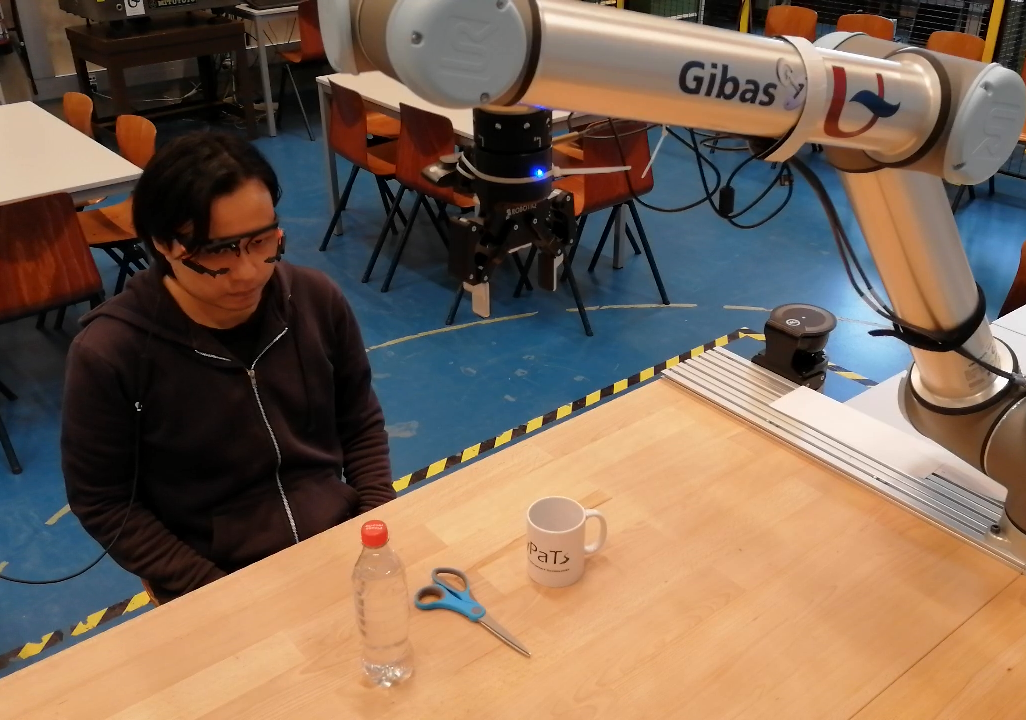}
\caption{The scenario where a human can select one of the objects on the table by eye tracking glasses. A robotic manipulator will pick up the selected object.}
\label{fig:scene}
\end{figure}

\begin{figure*}[htbp]
\centering
    \begin{subfigure}[t]{0.3\textwidth}
    \includegraphics[width=1.0\textwidth]{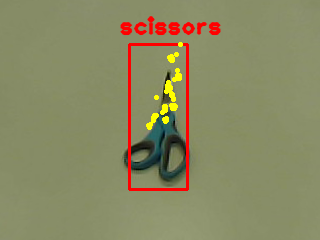}
    \caption{Gaze patterns of observation over time. Yellow dots are the gaze points.}
    \label{fig:fix_sac}
    \end{subfigure}
    \quad
    \begin{subfigure}[t]{0.6\textwidth}
    \includegraphics[width=0.5\textwidth]{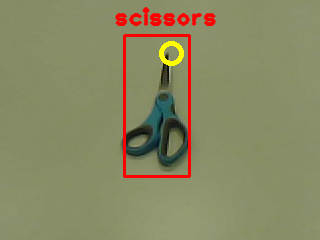}
    \includegraphics[width=0.5\textwidth]{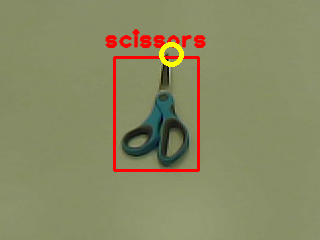}
    \caption{Gaze outside of bounding box caused by the variation of the bounding box. Yellow circle is a single gaze point.}
    \label{fig:out_bbox}
    \end{subfigure}
\caption{(a): The recorded gazes over time when a human is looking at the scissors. (b): Two consecutive frames during the observing period with tracked gaze and detected bounding box. The variation of the size of the bounding box comes from the detection algorithm and it can cause the gaze being outside of bounding box while the human intention is still on scissors.}
\label{fig:fixation_limitations}
\end{figure*}

In Human-Robot Interaction (HRI), fixation is often used as a model of the visual intention of a human. In \cite{yuan2019human}, when a fixation was classified, an image patch is cropped around the fixation point and it is fed to a neural network to detect a drone. In \cite{li20173} and \cite{wang2018free}, fixations were  both used for selecting an object to grasp. It was also used for selecting a grasping plane of an object in \cite{li20173}. However, there are  limitations in the use of fixation to select an object for further actions. Consider a scenario as displayed in Fig.\,\ref{fig:scene}. A human is wearing a mobile eye tracking device and he or she can select detected objects on the table by gaze. A robot then will pick up the object for him or her. Using fixation to select an object yields that a human has to look at a very small region of the object. When observing an object, the human gaze is hardly fixating on a small region. The authors in \cite{takahashi2018system} demonstrated that the human gazes spread over different regions of an object which was being observed. Fig.\,\ref{fig:fix_sac} shows one example, the human gazes when observing a pair of scissors are plotted with yellow dots. This implies that saccades occur during the process of observing. Thus, the use of fixation may suffer from loss of information in the sense that when a saccade is classified, the human intention is still on the object. Furthermore, given the detected object being represented by a bounding box, the center of a fixation may fall out of the bounding box while the human is still looking at the object. We refer it as gaze intention error. This error have different sources. First is the fluctuation of the size of bounding box (Fig.\,\ref{fig:out_bbox}) which is caused by the object detection algorithm. Second, a poor calibration would also result in this error. Moreover, the head mounted mobile eye tracking device may accidentally be moved after calibration and the detected gaze will be shifted. This shift is another source of the error.

We propose a novel saliency and Earth Mover's Distance (EMD) based approach to detect human intention which can overcome the limitations mentioned above. In computer vision, saliency is a Bottom-Up model of visual attention \cite{frintrop2015traditional}. A saliency map is an image which provides regions that a human could possibly look at. For a detected object, its salient regions can be extracted by computing the saliency map within the bounding box. The salient regions of an object can be interpreted as the hypotheses of where a human being's visual focus is located. The gaze signals from the mobile eye tracking device, on the other hand, can be interpreted as the actual location that a human is looking at. Hence we can form the problem of detecting which object human intention is located on from a different perspective. Unlike  detecting fixations and checking whether the fixation center is on objects, we calculate the similarity between the hypothetic gazes over objects and actual gazes. We convert the saliency maps of objects into the hypothetic gaze distribution and generate actual gaze distribution by collecting gaze information from the eye tracking glasses. EMD distance is used to compare the similarity between these two distributions\cite{rubner2000earth}. We further use 1-Nearest Neighbor (1-NN) with null rejection as classifier to determine which object is being observed. To interact with robot, we define a simple eye gesture to confirm that an object has been selected. Our results show that the proposed method can significantly increase the accuracy in predicting the human intention with the presence of saccades and gaze intention errors. Compared to using fixation, the percentage of the correct predicted intention increases around 18$\%$. For the interaction with the robot, our approach can reach 92.2$\%$ success rate within made predictions.

The rest part of the paper is organized as following: In Section II, we review the related work. In Section III, we explain our method in detail. Experimental results are shown in Section IV and we conclude our work in Section IV.

\section{Related Work}

Saliency computation is a Bottom-Up method to generate visual attention from the image data. Several biologically inspired models have been used for salient region detection \cite{itti1998model,frintrop2006vocus,montabone2010human, frintrop2015traditional}. All of these work are based on the cognitive model called Feature-Integration Theory of Attention\cite{treisman1980feature} which assumes the features in a scene (i.e. color, orientation, brightness, motion) is processed and registered in parallel by visual system. 
%The authors in \cite{itti1998model} used a biologically inspired model to compute saliency map. This model simulates the human visual system and cognitive function. 

In \cite{itti1998model}, images are fed to color, intensity and orientation channel in parallel. Image pyramids are created within the channels. Center-Surround operation is applied in each channel with image features to create feature maps. The Center-Surround operation simulates the function of ganglion cells in human retina \cite{frintrop2011computational}. Six feature maps are generated for intensity channel, 12 for color channel and 24 for orientation channel. A normalization operator is then applied on each channel which will suppress the homogeneous area in feature maps. A conspicuity map for each channel is obtained by applying cross scale addition after normalization. The final saliency map is computed by linearly combining the three conspicuity maps. VOCUS\cite{frintrop2006vocus} made improvements based on  \cite{itti1998model}. and it achieved better results with the cost of more computation time. VSF \cite{montabone2010human} calculated on-center and off-center differences on integral images with original image size instead of on scaled image, which can retain fine grained information without increase computational cost. VOCUS\cite{frintrop2006vocus} made improvements based on  \cite{itti1998model} and it achieved better results with the cost of more computation time. VSF \cite{montabone2010human} calculated on-center and off-center differences on integral images with original image size instead of on scaled image, which can retain fine grained information without increase computational cost.

More recently, various Deep Neural Network (DNN) based approaches were used for predicting the visual attention such as DeepGaze\cite{kummerer2014deep}, SALICON \cite{huang2015salicon}, DeepFix\cite{kruthiventi2017deepfix}. Theses networks use deep features pre-trained on datasets for image recognition and fine-tuned for saliency prediction\cite{kummerer2017understanding}. Despite the significant improvements of the performance on various datasets, the low level Bottom-Up methods are underestimated.\cite{bylinskii2016should,kummerer2017understanding,kong2018deep}.

EMD is a metric of two distributions which can be used to measure the similarity of two distributions. It was first introduced into computer vision field \cite{rubner2000earth} in \cite{peleg1989unified}. The EMD was distance also used in image retrieving\cite{rubner2000earth,bazan2019quantitative}. The information of histograms of images were derived to construct the signatures of images $P=\{(p_1, w_{p_1})...(p_m, w_{p_m})\}$ and $Q=\{(q_1, w_{q_1})...(q_n, w_{q_n})\}$ where $p_i$, $w_{p_i}$, $m$ and $q_j$, $w_{q_j}$, $n$ are the mean of cluster, weighting factor and number of clusters of the respective signature. Distance matrix $\mathbf{D}$ is the ground distance between $P$ and $Q$ and flow matrix $\mathbf{F}$ describes the cost of moving "mass" from $P$ to $Q$. EMD distance is the normalized optimal work for transferring the "mass". In \cite{bazan2019quantitative}, EMD is compared with other metrics, i.e. Histogram Intersection, Histogram Correlation, $\chi^2$ statistics, Bhattacharyya distance and Kullback-Leibler(KL) divergence, for measuring image dissimilarity in color space. EMD had better classification performance than the other metrics. It was also shown that EMD can avoid saturation and remain good linearity when the mean of target distribution changes linearly.

\section{Methodology}

Our methodology will be applied in the scenario described in Section \ref{sec:intro} and shown in Fig.\,\ref{fig:scene}. We use head-mounted eye tracking device which provides the world image $I_w$ and gaze point $g(x,y)$ where $x$ and $y$ are the coordinates in $I_w$.  There are three objects, cup, scissors and bottle, in the scene. All objects are placed on a table. A human can select one of the objects and a robotic manipulator will pick up the desired object. We first detect all the objects by feeding the world image $I_w$ to an object detector. Then we generate hypothetic gaze samples on the detected objects,  and we compare them with actual gazes obtained from the head-mounted eye tracking device. Finally, the similarity between the hypothetic gaze and actual gaze is used to classify if the human intention is on an object.

\subsection{Object Detection}

We use deep learning based object detector YOLOv2 \cite{redmon2016yolo9000} to detect the objects in our scene. It has the advantage of the real time capability. The convolutional neural network of YOLOv2 is trained on COCO dataset\cite{lin2014microsoft}. The detected object is represented in the form of a bounding box $B$ and a class label $c$.
%=[x_{max}, x_{min}, y_{max}, y_{min}, c]$, where 

\subsection{Saliency As Hypothetic Gaze Distribution And Actual Gaze Distribution}
The YOLO bounding box is transformed into $B=[ p(x,y),w,h,c ]$ where $p(x,y)$, $w$ and $h$ are the center, width and height of the bounding box respectively and $c$ is the class label. Image patches are cropped out from $I_w$ with the sizes of the bounding boxes of detected objects. The object saliency maps are computed from the image patches using the algorithm in \cite{montabone2010human}, then we sort the pixels of the saliency map by the intensity in descent order and take first $l$ pixels. From these $l$ pixels, $k$ pixels are randomly sampled following unit Gaussian distribution, which can be interpreted as hypothetic gaze points. Next we calculate the Euclidean distance between each of the $k$ pixels and the center of bounding box. This distance distribution is denoted as hypothetic gaze distribution $\pi_s$. To form the actual gaze distribution, we define a temporal window of size $k$, for each actual gaze points(acquired from eye tracking device) in the window, we also calculate its Euclidean distance to the center of bounding box, the resulting distance distribution is denoted as actual gaze distribution $\pi_g$.

%\begin{equation}
% v_p=[S^1_p(x,y), ... S^i_p(x,y), ... S^n_p(x,y)], n=w \times h 
%\label{eq:saliency_vector}
%\end{equation} 

%where $S^1_p(x,y)$ is the pixel with highest intensity in $S_p$. From first $l$ largest values in $v_p$, we take $k$ samples following a unit Gaussian distribution from $v_p$ and 

EMD is used as the measure of the similarity between distributions $\pi_s$ and $\pi_g$. In order to use EMD, the distributions need to be transformed into signatures. We first calculate the geometric distance histograms $H_s=\sum_{i=1}^mb_s^i$ for $\pi_s$ and $H_g=\sum_{j=1}^nb_g^j$ for $\pi_g$, where $m$ and $n$ are the number of bins. The signatures $\mathbf{s}_s$ and $\mathbf{s}_g$ are calculated similar to \cite{bazan2019quantitative},

\begin{equation}
%\begin{split}
 \mathbf{s}_s=\sum_{i=1}^mb_s^iw_s^i  , \quad
 \mathbf{s}_g=\sum_{j=1}^nb_g^jw_g^j
%\end{split}
\label{eq:sig}
\end{equation} 

where $b_s$ and $b_g$ are the bin values from $H_s$ and $H_g$ and weighting factors $w_s$ and $w_g$ are the middle values of the respective bin intervals. The distance matrix $\mathbf{D}_{sg}=[d_{ij}]$ is the ground distance between $\pi_s^i$ and $\pi_g^j$. The flow matrix $\mathbf{F}_{sg}=[f_{ij}]$ is the cost of moving the "mass" from $\pi_s$ and $\pi_g$. The work function is,

\begin{equation}
 Work(\mathbf{D}_{sg},\mathbf{F}_{sg})=\sum_{i=1}^m\sum_{j=1}^nd_{ij}f_{ij}.
\label{eq:work}
\end{equation}

The EMD distance is calculated as,

\begin{equation}
 EMD(\pi_{s},\pi_{g})=\frac{min( Work(\mathbf{D}_{sg},\mathbf{F}_{sg}))}{\sum_{i=1}^m\sum_{j=1}^nf_{ij}}.
\label{eq:emd}
\end{equation}

For classifying which object a human is focusing on, EMD distance is calculated as the measure of similarity between hypothetic gaze and actual gaze. Then we find 1-NN among all object signatures with null rejection. If the EMD value of the nearest object is below the rejection threshold, an empty class is assigned which means the human is not looking at any of the objects.

\subsection{Interaction With The Robot}
\label{subsec:interaction}

The robot needs to know which object is selected by the human to perform the  grasping task. Our approach provides the information of which object a human is looking at, but it does not necessarily mean that the object is also the one that the human wants the robot to pick up. Hence a simple eye gesture is defined to confirm the selection by the human. After looking at the intended object, closing the eyes for 0.5 seconds and opening the eyes again will let the robot know the selected object. For each object, a trajectory and a grasping profile is pre-defined for grasping. The robot will pick up the object selected and confirmed by the human participant.

\section{Experiment And Result}

We use Pupil Labs eye tracking glasses\cite{Kassner:2014:POS:2638728.2641695} for eye tracking. The YOLOv2 object detector is implemented in ROS\cite{bjelonicYolo2018}. We use an UR10 robot for grasping, the communication with it is also via ROS \cite{andersen2015optimizing}.

\subsection{Data Collection}

We have asked 9 people to participate in the experiments. All participants are aged in the range of 20 to 40, all of them have research background in engineering. One of the participant has experience in eye tracking. The rest has no prior experience in eye tracking.

\subsection{Experiments}

\subsubsection{Observing Objects}

In this experiment, the participants are asked to observe the object. There is only one object is placed on the table at a time. The participants are asked to observe the object first and then look away from the object. The participant can freely look in the scene during the "look away" period. The procedure repeats until all objects are observed individually. All blinks in the experiments are not considered.

We compare EMD similarity measure with KL divergence and Bhattacharyya distance between histogram $H_s$ and $H_g$. Our approach will also be compared with fixation based approach. We use I-DT implemented by Pupil Labs for fixation detection, the rest events are considered as saccade events. For every detected fixation event, the human intention is on the object if the fixation center is inside the bounding box of the object.

\subsubsection{Interaction}

In this experiment, the participants are asked to select the one of the objects on the table using the interaction method described in Section \ref{subsec:interaction}. After one selection is made, an audio feedback will be given to the participant informing the prediction of the selection. Then the participant can start to make next selection. Before the experiments start, the participants have two minutes to practice the eye gesture and get familiar with the object selection process. During the experiment, each participant keeps selecting object until 10 successful predictions are made. 

\subsection{Evaluation}
For the "Observing Objects" experiment, we use Cohen's Kappa to evaluate sample-to-sample accuracy instead of other commonly used metrics such as Precision-Recall and F1 score. Cohen's Kappa measures the agreement between two sets of data. The value 0 means no agreement and value 1 means perfect agreement. Using Cohen's Kappa to compare the predictions and ground truth will give the result how they agree with each other and it can be interpreted as accuracy. When evaluating imbalanced data, Cohen's Kappa is a better option than Precision-Recall and F1 score. In our experiment, we compare our method with fixation based method. The experimental data is classified into fixations and saccades. The majority of the data belongs to fixation events, thus using Cohen's Kappa will give a better understanding of the results. We further perform a event analysis using the metrics similar to \cite{steil2018fixation}. The correct events and deletion events are evaluated. When a predicted event equals to the ground truth, it is considered as a correct event. A deletion event is the event missed by prediction but exists in ground truth. The deletion events are categorized  into three types, the deletion caused by algorithm, the deletion caused by gaze intention error and deletion caused by saccade. A deletion event caused by gaze intention error can be determined by checking if the gaze point locates outside of the object bounding box. Although our approach doesn't classify fixation events nor saccade events, the saccadic eye movement would affect the classification score. By comparing saccade events with the detected events using our approach, we can still know if a deletion is caused by saccade. For the fixation based method, the deletion by algorithm doesn't apply since the algorithm only checks if fixation point is inside bounding box, it is same as the deletion by gaze intention error.

For the "Interaction" experiment, we also analyze the correct and deletion events. A correct event is defined as, after a participant makes object selection by eye gesture, the predicted object is the object the participant wants to select. A deletion event is defined as, the predicted object is different from the object that the participant wants to select by eye gesture. There are two kinds of deletion events, deletion caused by misclassification and deletion caused by missed detection. A misclassification causes one deletion event and one insertion event, since the correct object is not predicted and a wrong object is in the prediction. The insertion event is, however, correlated to the deletion by misclassification, thus we don't evaluate the insertion event. A deletion caused by missed detection is defined as, a participant use eye gesture to select an object but no prediction is made due to the failure of recognizing the eye gesture. 

\subsection{Results}

\begin{figure}[t]
\centering
\includegraphics[width=0.5\textwidth]{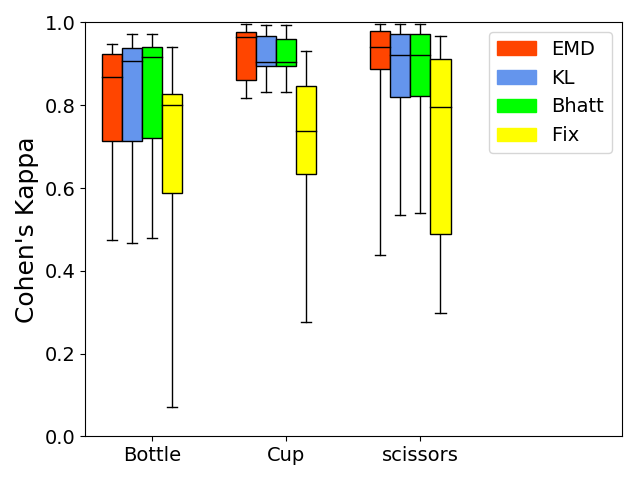}
\caption{The result of all participants in experiment "observing objects". The plot shows Cohen's Kappa for each observed object with our approach (EMD), similarity measure with KL divergence (KL) and Bhattacharyya distance (Bhatt), and with fixation approach (Fix). }
\label{fig:fix_emd_reuslt}
\end{figure}

\begin{figure*}[t]
\centering
\begin{subfigure}[t]{0.329\textwidth}
	\includegraphics[width=1.0\textwidth]{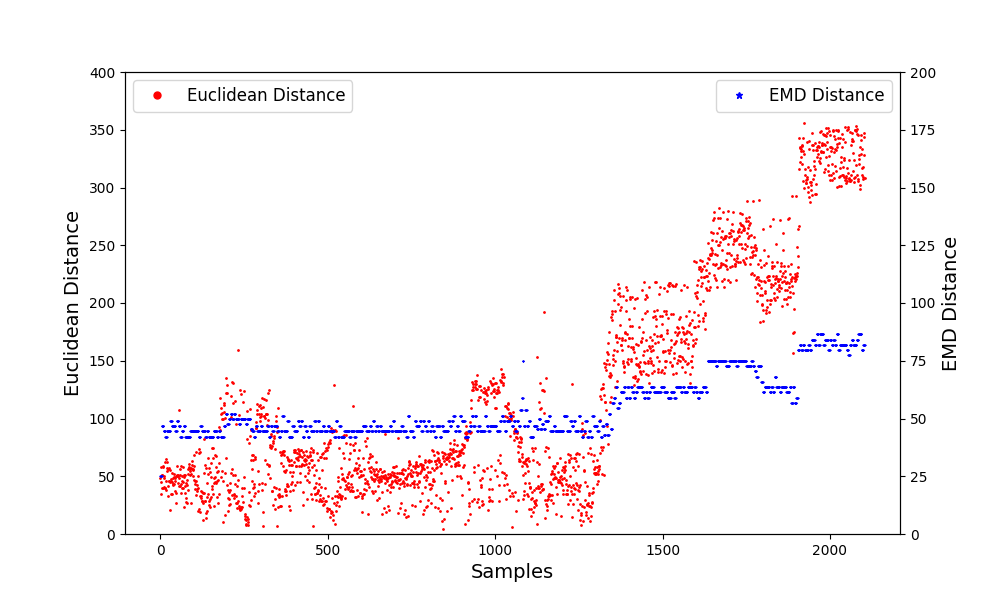}
	\includegraphics[width=1.0\textwidth]{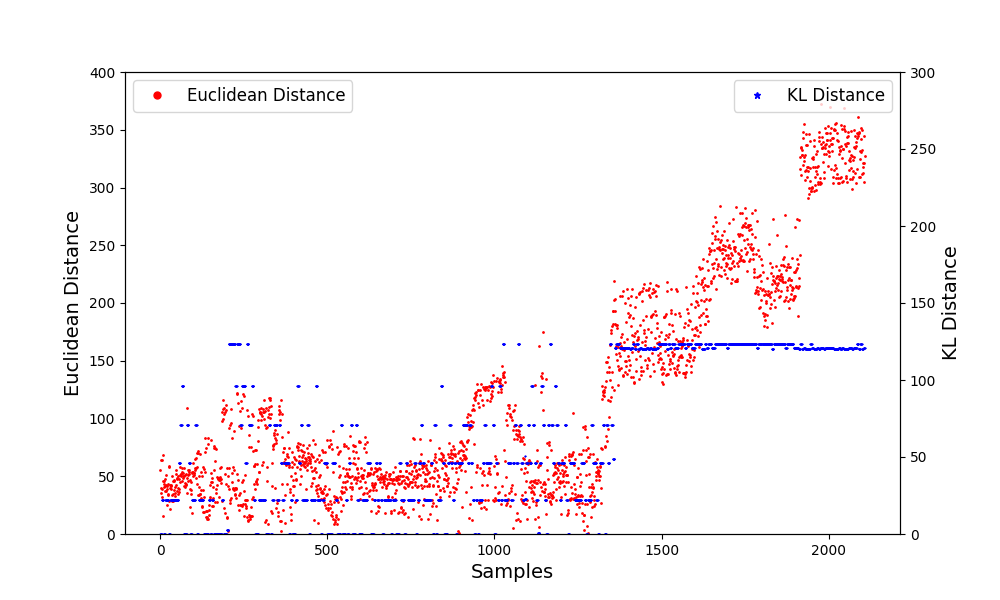}
	\includegraphics[width=1.0\textwidth]{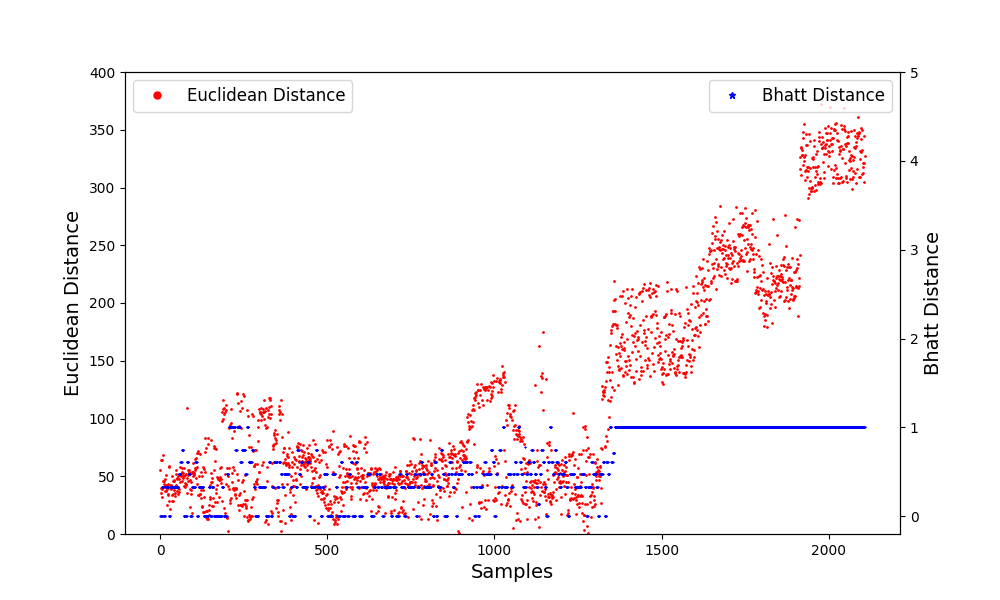}
	\caption{Bottle}
    \label{fig:linear_dist_bottle}
    \end{subfigure}
\begin{subfigure}[t]{0.329\textwidth}
    \includegraphics[width=1.0\textwidth]{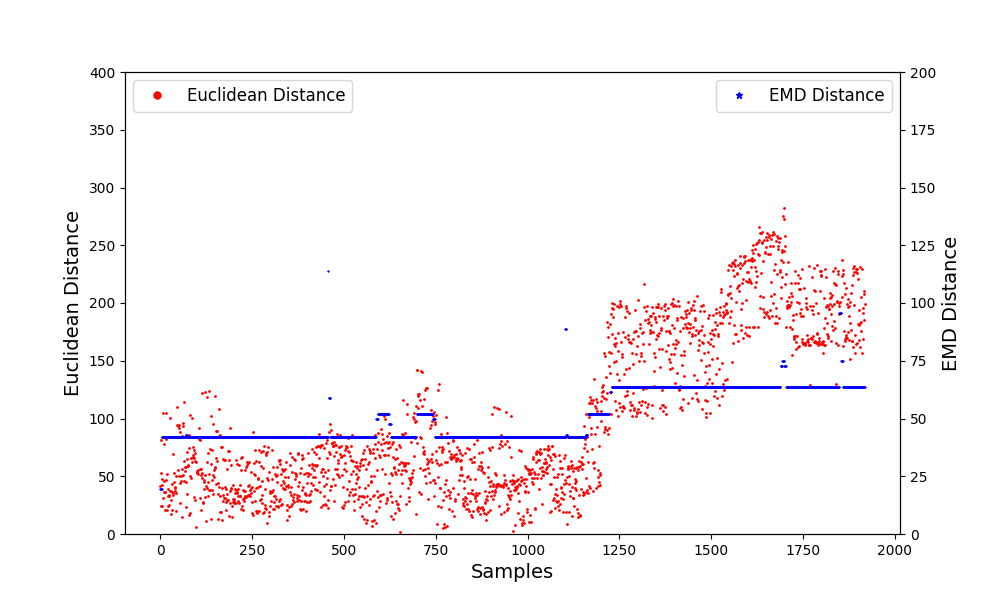}
    \includegraphics[width=1.0\textwidth]{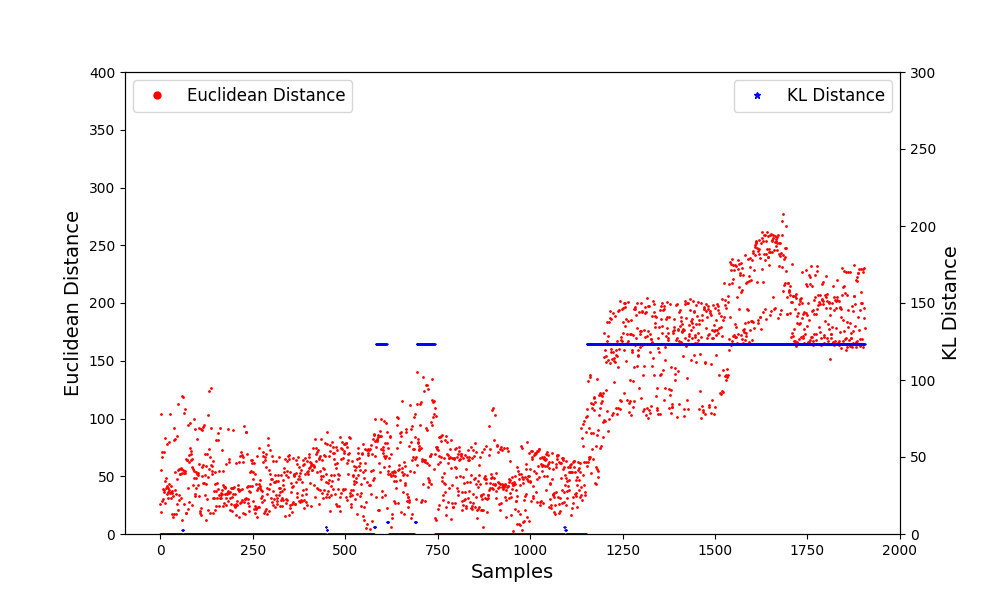}
    \includegraphics[width=1.0\textwidth]{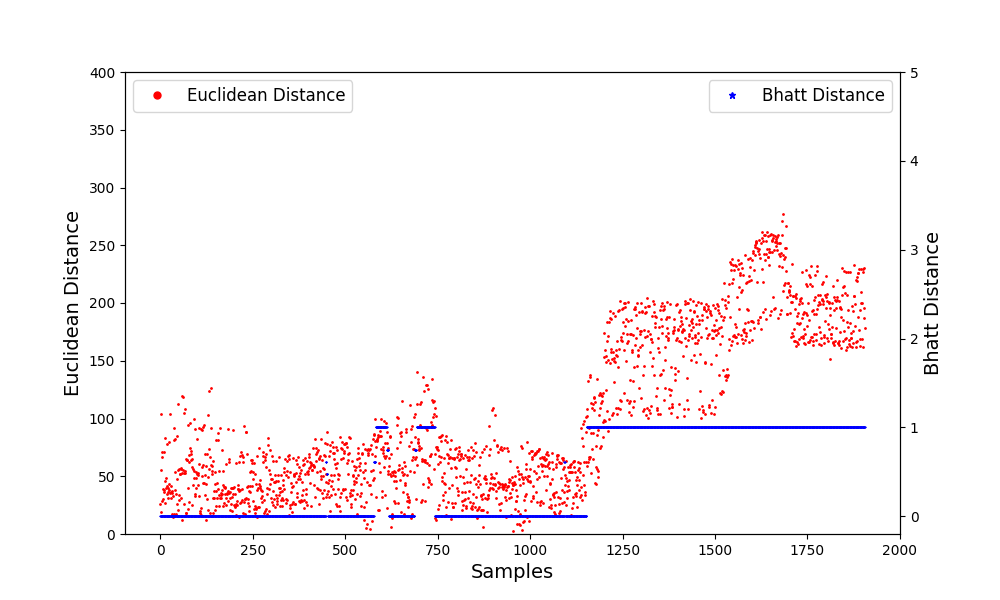}
    \caption{Cup}
    \label{fig:linear_dist_cup}
    \end{subfigure}
\begin{subfigure}[t]{0.329\textwidth}
    \includegraphics[width=1.0\textwidth]{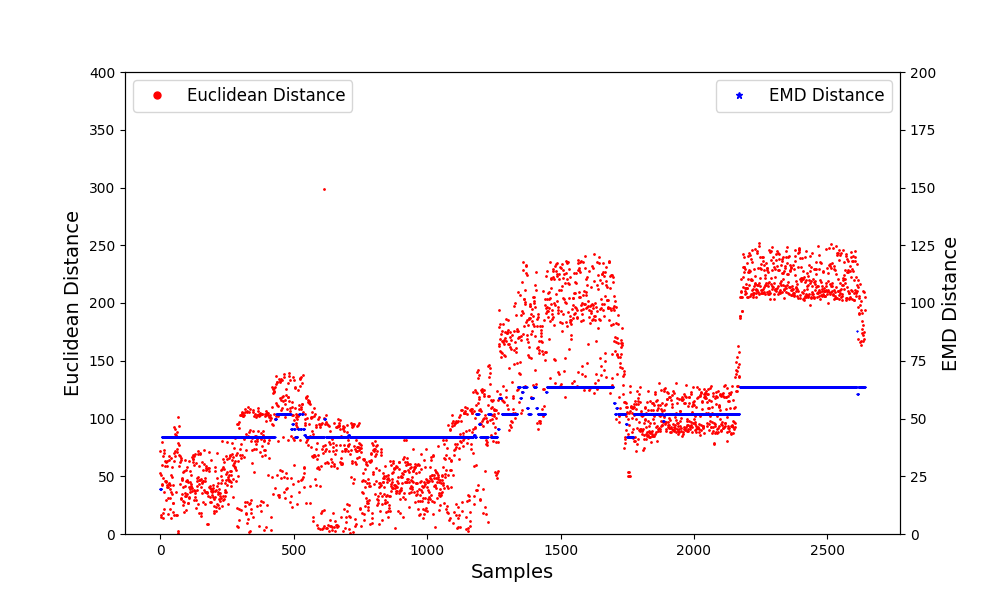}
    \includegraphics[width=1.0\textwidth]{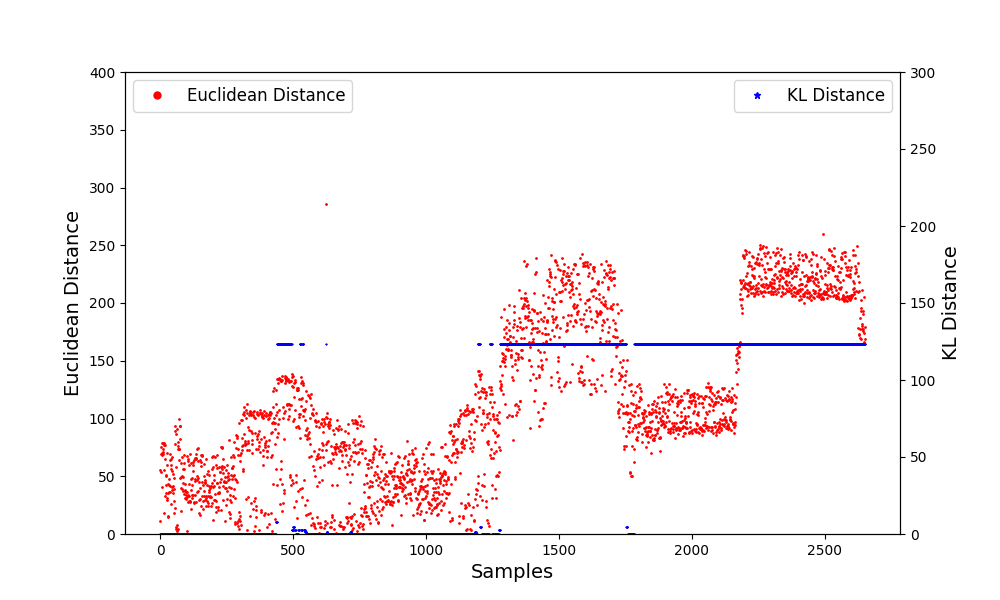}
    \includegraphics[width=1.0\textwidth]{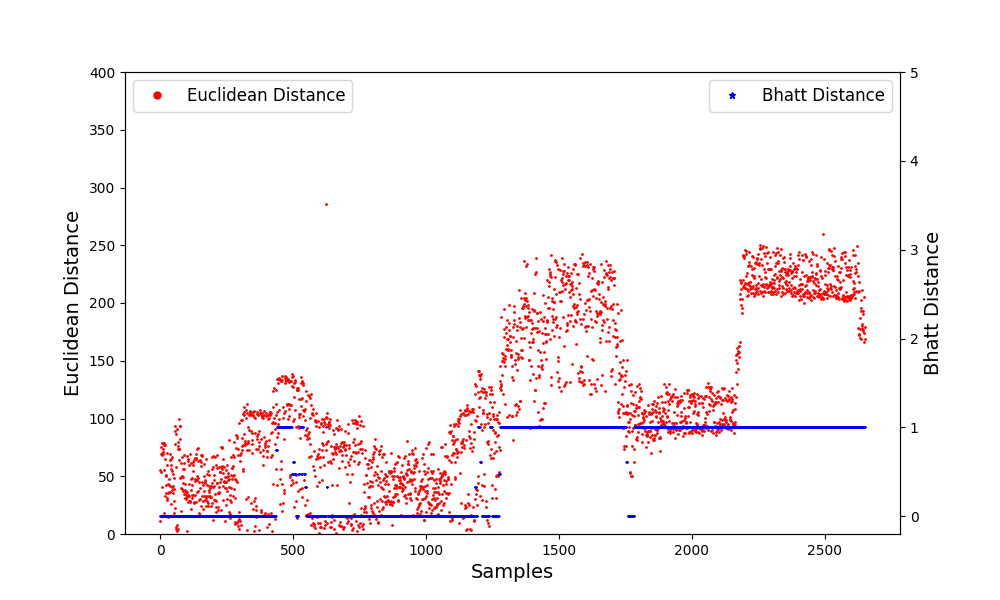}
    \caption{Scissors}
    \label{fig:linear_dist_scissors}
    \end{subfigure}
\caption{Distances plots of one participant in experiment "Observing Objects". The similarity distances (EMD, KL and Bhattacharyya) are the distances between saliency distributions and gaze distributions. Euclidean distance calculates the distance between actual gaze points and hypothetic gaze points extracted from saliency map of objects.}
\label{fig:linear_dist}
\end{figure*}

Fig.\,\ref{fig:fix_emd_reuslt} shows the Cohen's Kappa of sample-to-sample classification in the experiment "Observing Objects". The plots summarize the results of our approach, KL divergence and Bhattacharyya distance as similarity measure and with fixation based method. For the three metrics of similarity measure, EMD, KL divergence and Bhattacharyya distance, they have comparable results in all three objects. But as pointed in \cite{bazan2019quantitative}, the KL divergence and Bhattacharyya distance will quickly saturated when the difference of two distributions increases. The authors tested on artificial data, but it is also valid in our real experiments. Fig.\,\ref{fig:linear_dist} displays the similarity distances and Euclidean distances of all three objects from one participant. The similarity distances are the EMD, KL divergence and Bhattacharyya distance between the actual gaze distribution and the hypothetic gaze distribution. The Euclidean distances are the geometric distance between actual gaze points and the hypothetic gaze points. Take the bottle case as an example, the first 1302 events is the period when the participant is observing the object and afterwards is the period of looking away from the object. The step like changes in the plots mean the gazes are changed from one region to another. In the "look away" period, the EMD distance can follow the trend of the step changes while KL divergence and Bhattacharyya distance reach their limits. Even in the period when the participant is observing the object, KL divergence and Bhattacharyya distance already are already saturated in a few samples. The saturations can be observed in cup and scissors cases in Fig.\,\ref{fig:linear_dist_cup} and Fig.\,\ref{fig:linear_dist_scissors} too.

%\begin{table}[t]
%\renewcommand{\arraystretch}{1.5}
%\caption{Correct Events And Deletion Events In Observing Objects Experiment}
%\label{table:event_fix_sac}
%\centering
%\resizebox{\linewidth}{!}{
%\begin{tabular}{c c||c c  c c}
%\hline
%& &\multicolumn{4}{|c}{\bfseries Observing Objects} \\
%& & Correct[$\%$] & Deletion(algo)[$\%$] & Deletion(bbox)[$\%$] &Deletion(sac)[$\%$]  \\
%\hline\hline
%\multirow{2}{*}{\bfseries Bottle} & EMD & 93.5 & 2.7 & 3.2 & 0.6 \\
%&Fixation & 73.7 & n/a & 20.7 & 5.6\\
%\hline
%\multirow{2}{*}{\bfseries Cup} & EMD & 96.7 & 0 & 2.8 & 0.5 \\
%&Fixation & 75.5 & n/a & 18 & 6.5\\
%\hline
%\multirow{2}{*}{\bfseries Scissors} & EMD & 97 & 0.2 & 2.5 & 0.3 \\
%&Fixation & 83 & n/a & 12.2 & 4.8\\
%\hline
%\end{tabular}
%}
%\end{table}

\begin{table}[t]
\renewcommand{\arraystretch}{1.5}
\caption{Correct Events And Deletion Events In Observing Objects Experiment}
\label{table:event_fix_sac}
\centering
\resizebox{\linewidth}{!}{
\begin{tabular}{c || c c |c c |c c}
\hline
& \multicolumn{2}{c|}{\bfseries Bottle} & \multicolumn{2}{c|}{\bfseries Cup} & \multicolumn{2}{c}{\bfseries Scissors}\\
& EMD & Fix & EMD & Fix & EMD & Fix \\
\hline\hline
Correct[$\%$] & 93.5 & 73.7 & 96.7 & 75.5 & 97 & 83 \\
\hline
Deletion(algo)[$\%$] & 2.7 & n/a & 0 & n/a & 0.2 & n/a \\
\hline
Deletion(bbox)[$\%$] & 3.2 & 20.7 & 2.8 & 18 & 2.5 & 12.2 \\
\hline
Deletion(sac)[$\%$] & 0.6 & 5.6 & 0.5 & 6.5 & 0.3 & 4.8 \\
\hline
%& &\multicolumn{4}{|c}{\bfseries Observing Objects} \\
%& & Correct[$\%$] & Deletion(algo)[$\%$] & Deletion(bbox)[$\%$] &Deletion(sac)[$\%$]  \\
%\hline\hline
%\multirow{2}{*}{\bfseries Bottle} & EMD & 93.5 & 2.7 & 3.2 & 0.6 \\
%&Fixation & 73.7 & n/a & 20.7 & 5.6\\
%\hline
%\multirow{2}{*}{\bfseries Cup} & EMD & 96.7 & 0 & 2.8 & 0.5 \\
%&Fixation & 75.5 & n/a & 18 & 6.5\\
%\hline
%\multirow{2}{*}{\bfseries Scissors} & EMD & 97 & 0.2 & 2.5 & 0.3 \\
%&Fixation & 83 & n/a & 12.2 & 4.8\\
%\hline
\end{tabular}
}
\end{table}

The Kappa score of fixation in general is less than any of the similarity measure. It is interesting to notice that the fixation score is also more uncertain. The highest Kappa scores of all objects are over 0.9. The lowest ones are 0.07, 0.28 and 0.36 respectively. The reason of this large variation is due to the different cleanliness of the data from the participants. The data is more clean if it contains less saccades and less gaze points which are located outside the bounding boxes. It is very intuitive that the more clean data will result in higher score in fixation method and the less clean data will produce lower score. For instance in the bottle case, the majority of the gazes of one participant is located out of the area of the bounding box, thus the kappa of fixation method is 0.7. In the bottle and scissors case, the lowest scores of similarity measures are much lower than the rest ones. In these sets of data, during the periods when participants look away from the objects, the gazes are still close to the objects. This means when the participants look at the places very close to the objects, our approach can not correctly classify whether the human intention is on object or not.

%The deletion events are further divided into the deletion caused by saccade and deletion caused by gaze intention error.

Table \ref{table:event_fix_sac} provides more insight in how saccade events and gaze intention errors affect the accuracy. We collect the events while the participants are observing the objects. The number of correct events and deletion events of all participants are summed together per object. The total number of detected events in bottle, cup and scissors using our approach are 2336, 2206, 2450, and 1998, 2898, 2055 for the fixation based method. The results indicates that our approach can significantly improve the performance. The percentages of deletion events with regard to all events in bottle, cup and scissors using our approach are 6.5$\%$, 3.3$\%$ and 3.0$\%$. And for fixation method, the respective percentages are 26.3$\%$, 24.5$\%$ and 17.0$\%$. In addition, the number of the deletion caused by gaze intention error are reduced and the deletion caused by saccade can almost be eliminated. 

%The EMD distance is also linearly correlated to the Euclidean distance between the actual gaze points and hypothetic gaze points.

Table \ref{table:selection} gives the result of the "interaction" experiment. The events numbers are the sum of events of all participants. In total, 119 attempts are made to select an object in the scene. A fairly high portion of attempts (36) are deletion events, among which 29 events are deleted due to the failure of giving prediction and 7 are due to misclassification. Since our interaction eye gesture involves in closing eyes for 0.5 seconds, it is difficult for the human participants to accurately measure the time durations. This causes the failure of recognizing eye gesture. However, within the successful predictions (correct event plus deletion caused by misclassification), 92.2$\%$ of the selections are correctly predicted by our approach.

\begin{table}[t]
\renewcommand{\arraystretch}{1.5}
\caption{Correct Events And Deletion Events In Interaction Experiment}
\label{table:selection}
\centering
\resizebox{\linewidth}{!}{
\begin{tabular}{c|| c }
\hline
%\multicolumn{2}{c}{\bfseries Interaction } \\
& Prediction Event \\
\hline\hline
Correct[$\#$] &  83\\
\hline
Deletion(classification)[$\#$] & 7\\
\hline
Deletion(detection)[$\#$] & 29\\
 %Prediction Event & 83  & 7 & 29 \\
 %Prediction Event & 69.7 & 5.9 & 24.4 \\
\hline
\end{tabular}
}
\end{table}

\section{Conclusion}

In this work, we propose a new approach to determine the intention in gaze-based Human-Robot Interaction. The interaction scenario is that a human can select one of the objects in the scene by gaze and a robot can pick up the selected object. To determine which object the human is looking at, we calculate the similarity between the hypothetic gaze points on the objects and the actual gaze points and use 1-NN to classify the intended object. The hypothetic gaze points are sampled from the salient region calculated by a a visual cognition model. We use EMD distance to measure the similarity between hypothetic gaze distribution and actual gaze distribution. 

In our experiments, we compared EMD distance with KL divergence and Bhattacharyya distance. Although EMD distance has comparable results with the other two, it doesn't suffer from the saturation and it has better correlation to the Euclidean distances between actual gaze and hypothetic gaze. We also compared our approach with fixation based approach, results showed that our approach can significantly restrain the deletion caused by the gaze intention error and saccade. We also demonstrated the high performance in interacting with robot. Although a high percentage of attempts in selecting object is failed, it is due to the difficulty in controlling the time of eye closing. For the successful predictions, 92$\%$ of them are correctly predicted. The failed predictions could be improved when the participants get more familiar with the eye gesture for confirmation. The main limitation of our approach is that, it can not cognitively classify that the human is not focusing on the object but looking at the places very close to the object. 

Overall, in predicting if human is look at an object in a HRI scenario, our approach has better performance than the fixation in two aspects. First, our approach is almost not effected by the saccades when a human is observing different parts of an object. Second, our approach is more robust to the gaze intention error. When the gaze data is clean(i.e. gaze points is inside object bounding box), using fixation can have comparable results. But when the data contains nosies so that gazes are located outside of bounding box(caused by variation of bounding box, calibration, etc), our approach can still accurately predict the human intention while fixation method will fail. 

\makeatletter
\def\endthebibliography{%
  \def\@noitemerr{\@latex@warning{Empty `thebibliography' environment}}%
  \endlist
}
\makeatother
\bibliographystyle{IEEEtran}
\bibliography{ref.bib}

\end{document}